\documentclass[twoside,11pt]{article}

\usepackage[letterpaper, margin=1in]{geometry}
\usepackage{natbib}
\usepackage{microtype}
\usepackage{graphicx}
\usepackage{wrapfig}
\usepackage{mdframed}
\usepackage{subcaption}
\usepackage{hyperref}
\usepackage{xcolor}
\usepackage{amsmath}
\usepackage{amssymb}
\usepackage{mathtools}
\usepackage{enumitem}
\usepackage{csquotes}
\usepackage[normalem]{ulem}
\usepackage{booktabs}
\usepackage{bm}

\setlist[itemize]{labelindent=16pt}
\setlist[enumerate]{labelindent=16pt}
\setlist[description]{font=\normalfont\space,labelindent=16pt,leftmargin=48pt}

\begin{document}

\title{Towards autonomous artificial agents with an active self:\\modeling sense of control in situated action\footnote{This is a pre-print of a manuscript that has been accepted for publication at Cognitive Systems Research. Please always cite the peer-reviewed article under \url{https://doi.org/10.1016/j.cogsys.2021.11.005}}}
\author{Sebastian Kahl\textsuperscript{a}\footnote{Corresponding author: \texttt{skahl@uni-bielefeld.de}}~, Sebastian Wiese\textsuperscript{b}, Nele Russwinkel\textsuperscript{b}, Stefan Kopp\textsuperscript{a}\\
\\
\textsuperscript{a}Social Cognitive Systems Group, CITEC, Bielefeld University,\\Bielefeld, Germany\\
\textsuperscript{b}Kognitive Modellierung in dynamischen Mensch-Maschine-Systemen, TU-Berlin,\\Berlin, Germany
}

\date{\vspace{-2ex}}
\maketitle

\begin{abstract}
In this paper we present a computational modeling account of an active self in artificial agents. In particular we focus on how an agent can be equipped with a sense of control and how it arises in autonomous situated action and, in turn, influences action control. We argue that this requires laying out an embodied cognitive model that combines bottom-up processes (sensorimotor learning and fine-grained adaptation of control) with top-down processes (cognitive processes for strategy selection and decision-making). We present such a conceptual computational architecture based on principles of predictive processing and free energy minimization. Using this general model, we describe how a sense of control can form across the levels of a control hierarchy and how this can support action control in an unpredictable environment. We present an implementation of this model as well as first evaluations in a simulated task scenario, in which an autonomous agent has to cope with un-/predictable situations and experiences corresponding sense of control. We explore different model parameter settings that lead to different ways of combining low-level and high-level action control. The results show the importance of appropriately weighting information in situations where the need for low/high-level action control varies and they demonstrate how the sense of control can facilitate this.
\end{abstract}


\section{Introduction}
Humans can learn new skills such as, e.g., inline skating through a process of gaining more and more control over how their actions will yield specific consequences. At first, it is necessary to put full attention to achieve a movement toward a goal a few meters away. After some training, one is able to skate and at the same time focus on a conversation with a friend. Even when going on coarse ground – with a lot of noise in haptic feedback – the skater is still confident that she has control over the situation. And, after taking off the skates, she will immediately switch back to the “normal-no-wheel” state of movement control. 

This example demonstrates that autonomous embodied agents need to be able to adapt how they control their situated action based on the current state of their body in a possibly uncertain environment and, importantly, an assessment of how they feel ``in control'' of acting in it. In the previous example, when skating over an uneven surface, movements may become unreliable but an experienced skater will easily be able to compensate for this without switching to another strategy of movement control (e.g., going slowly or starting to walk). However, when not being able anymore to keep up a reliable or stable control, she will decide to change the mode of action control. Underlying this remarkable ability must be some form of representation of oneself as an entity acting in and physically coupled with the world. And this ``self-representation'' must be involved in the interplay of bottom-up processes (sensorimotor learning and fine-grained adaptation of control) with top-down processes (cognitive processes for learning, planning, and decision making).

One property of embodied cognitive systems emerging from this interplay of perception-action processes, and being essential during action learning and control, has been termed ``active self''. It is commonly conceived of as a concept related to one’s phenomenal experience when acting in the here and now, and to the question of how we perceive ourselves to be in a particular situation. The active self has recently drawn a lot of interest of researchers in Psychology, Neuroscience, Philosophy, and Robotics.
For example, it has been suggested that the human cognitive system constantly updates and maintains a sense of ownership and agency about itself when acting in a dynamic spatial environment - key components of the so-called ``minimal self'' \citep{Gallagher:2000,Frith:2000ksa} that has been described as a phenomenological experience which is pre-reflectively implicit in action \citep{gallagherSimulationTrouble2007} and on which higher-order conceptual attributions depend. More specifically, the sense of agency can be seen as a compound of several basic experiences which are, e.g., intentional causation, the sense of initiation, and the sense of control \citep{pacherie:ijn_00352565}. 
However, the detailed relations between these multiple conjectured concepts and, in particular, the specific possibly underlying cognitive processes are still not clear.

We posit that the concept of an active self is highly relevant also for artificial embodied systems such as robots. On the one hand, the construction and study of technical systems able to exert autonomous (i.e., self-driven and self-controlled) actions in a spatial environment provides a unique test bed for investigating the mechanisms of self and action control in computational accounts. This constructivist approach is prevalent in the field of developmental robotics and aims to contribute to basic research through synthesizing behaviors in artificial systems. On the other hand, situated autonomous systems such as mobile robots need to have a ``sense'' of what changes they can bring (or have brought) about in the environment in which they are set to act, with the bodily resources they are equipped with. This is necessary for them to act adaptively in uncertain or changing environments and to be reliable, explainable, and acceptable to humans envisioned to work with them side-by-side.

However, it is currently not clear what the cognitive and bodily mechanisms responsible for the emergence of an active self are, both in humans and in artificial agents. How does a sense of control develop in a situation, how is it represented, and what is its role in creating expectations and decisions in action control? For example, how does our self-assessment of control influence our actions in less predictable environments, and how do we adapt action patterns to new self-representations, e.g., when navigating with a big backpack through a crowded subway? Finally, how can we computationally model the required interplay of cognitive and embodied processes of action control, motor learning, and multi-sensory perception in technical artifacts?

In this paper, we present first steps towards answering these questions by developing an approach to model cognitive systems that possess and exploit a sense of control. We start by discussing related work in Psychology, Neuroscience, A.I. and Robotics and we point out open challenges and requirements that modeling attempts have yet to meet to enable goal-directed, adaptive action control in non-idealized (i.e., uncertain and unpredictable) environments. We will argue that a evaluation of the current and the prospective degree of control is a key element of such systems, and that these  action evaluations that integrate high-level contextual information with low-level sensorimotor aspects can be understood in terms of prediction-based processing. Specifically, we will adopt the \textit{free-energy principle} (FEP) \citep{Friston:2009iz} to describe these processes as prediction-error minimization through adaptation processes in perception and action.
In section \ref{sec:model} we will present a model of a general control architecture that accounts for those embodied cognitive mechanisms and their interplay during adaptive action control. This computational architecture builds on a broad cognitive modeling approach integrating concepts such as motor action, perception, and action control in an integrated model of top-down and bottom-up processing across two layers, a sensorimotor control layer (SCL) and a cognitive control layer (CCL). Crucially, it lays down structures and mechanisms to model a sense of control, its plasticity, and its role in action control. 
We then present a concrete task scenario and show how the model is implemented and applied in this scenario in order to evaluate it in simulation studies. After this we report and discuss results from such an evaluation of different model variants (parameter settings) applied in differently challenging control situations.
Finally, we will discuss the benefits of the integrated modeling approach proposed here and point out prospects and implications for future work in Cognitive Systems, A.I., and Robotics.

\section{Background and related work}
\subsection{Active self and sense of control}
The notion of the \textit{active self} is closely related to several other concepts put forward in Psychology, Cognitive Science, Neuroscience, or Robotics in order to conceptualize the mechanisms and experiences of embodied cognitive agents when acting in the world. These concepts include, among others, embodied self, sense of agency, sense/feeling of control, or sense of ownership. They are all tied to the embedded and embodied cognition underlying situated action through some bodily means.

We define the \textit{active self} to be the entirety of representations and processes that a cognitive agent holds at a certain point in time about its own affordances and significancies vis-a-vis its perceptions and actions in the world. 
That is, it is involved with the agent's assessment of, e.g., which actions it can fulfill, whether it brought about or can bring about a specific situational change, or what its current bodily means are. 
The active self is hence highly malleable and rooted in interactions with the world or current bodily conditions.
This has been thoroughly described, e.g., in rubber hand illusion experiments where the ownership of a subject's hand was manipulated by supplying visual and sensory signals for a rubber hand that correlates to the subject's perception of signals from the subject's real hand, but hidden from view \citep{botvinickRubberHandsFeel1998}.
It was shown that this manipulation not only influences the perceived source of a touch, i.e., coming from the rubber hand.
Also, the neural activity in the premotor cortex reflects the subject's feeling of ownership of the hand \citep{ehrssonThatMyHand2004}.

Likewise, \cite{prinzEmergingSelvesRepresentational2003} discusses ``emerging selves'' that are continuously constructed to serve specific purposes. 
We consider the active self to be one of these emerging selves, serving the purpose of informed goal-setting and control in situated action. 
In this perspective, the active self does \textit{not} include the agent's beliefs about its identity, self-image, personality, emotional state, etc. that are commonly assumed to constitute a so-called ``narrative self'' \citep{Gallagher:2000,dennettSelfCenterNarrative1992}.
Rather, the active self is suggested to consist of, at least, two components: a \emph{sense of agency} and a \emph{sense of (body) ownership} \citep{satoIllusionSenseSelfagency2005} with no or only minimal overlap (e.g., \cite{kalckertMovingRubberHand2012a}).
Generally, a person's \textit{sense of agency (SoA)} refers to the subjective awareness of being able to initiate and successfully execute (i.e., cause) one's volitional actions. It is assumed to arise from both predictive and postdictive (inferential) processes, the former of which making use of an ability to anticipate sensory consequences of own actions in order to detect deviations or to attenuate the intensity of incoming signals. 
One account to model these processes is based on the ``comparator model'', which posits a continuous comparison of predicted and perceived sensory consequences and accounts for disorders of awareness in the motor system or delusion of control \citep{Frith:2000ksa}. Postdictive processes, in contrast, involve inferences drawn after the action and based on higher-level causal beliefs, in order to determine the cause of observed events or to check whether they are contingent and consistent with specific intentions \citep{Wegner:1999wy}.

While the sense of agency is certainly a core ingredient of the active self, it is also a very broad and multi-faceted phenomenon that links up situated perception and action with causal inferences and world knowledge. We focus here on a narrower sub-aspect more confined to situated perception and action, which we nevertheless deem crucial for the active self: the \textit{sense of control} (SoC) in situated action. According to \cite{pacherie:ijn_00352565}, the sense of control refers to the extent to which one subjectively feels in control of a specific action. It can range from, on the one hand, everything happening exactly as intended and the agent feeling in full control of his action, to, on the other hand, everything going astray and the agent feeling completely ``powerless''. How this feeling comes about is presumably based on predictive and postdictive processes of perception and action control.
Further, it has been shown to originate from a temporal integration of evaluations of single acts over whole sequences of actions, so that, e.g., the fluency of action selection strongly impacts the sense of control \citep{Chambon:2012hu}. 
Accordingly, we assume that the sense of control does not only rest on bottom-up situational evaluation of the momentary success in motor control, but also includes broader contextual aspects and inferences through postdictive processes.

Generally, situated action control in humans is seen not as a unitary, centralized process, but to involve the interplay of different and distributed levels of action regulation (e.g., \cite{Badre:2018bw}) that span
multiple levels of a \textit{sensorimotor hierarchy}.
The production and control of motor action starts from a goal (action intention) which is detailed into complex motor programs that in turn steer lower control primitives \citep{Grafton:2007jf}. 
Likewise, recognition is often assumed to run mainly bottom-up, starting from lower-level features that are integrated spatially and temporally into more complex patterns to ultimately yield hypotheses about the underlying action goals or intentions. 
Crucially, representations for perception and action seem to be closely integrated and mutual influences and interferences have been discussed in theories of common-coding or perception-action coupling \citep{Prinz:1997uu,Kilner:2003jy,Chaminade:2005cq,James:2009bq,Sacheli:2012bb}. 

Central to hierarchical models of action is the differentiation of intentions of action into subtypes, with abstractions over simple movements that range from simple motor intentions to communicative intentions. By that they also break down the subjective experience of action \citep{Pacherie:2008dl,Chambon:2011ek}.
Our focus on the sense of control and the necessary hierarchical control architecture covers a subset of these  subtypes of action intentions that is limited to simple motor intentions for basic motor guidance and control, as well as superordinate intentions as a first abstraction over mere movements that also requires a situational anchoring (similar to M-intentions and P-intentions\\ \citep{Pacherie:2008dl}).

Related to this, perception and action are closely tied together in steering actions in a complex, dynamic environment towards a desired goal. This requires different levels of \textit{regulatory control}, from fully automated (unaware) to fully conscious (aware), depending on the available level of expertise and routinization \citep{hackerInsightIllusionThemes1986}. Recent research has shown that cognitive processes related to the monitoring and control of one's own mental states or cognitive processing, are used to influence behavior  \citep{boldtDistinctOverlappingNeural2020}. 
Importantly, increasing motor expertise not only automates a well-known action routine, but it also enables better predictions of one’s own as well as somebody else’s action outcomes \citep{Stapel:2016ek}. This is also an aspect of inter-personal variability with regard to an individual's subjective experience of action. People experience some actions more proximal by focusing on the details of an action (or \textit{how} an action is performed), while others experience it more with respect to distal consequences or implications (or \textit{why} an action is performed) \citep{vallacherLevelsPersonalAgency1989}.

Accordingly, \cite{pacherie:ijn_00352565} suggests several basic experiences of the sense of being in control: the sense of motor control, the sense of situational control, and the sense of rational control. 
The sense of motor control is mainly aware if (and that) something is wrong, whereas the other are also aware of \textit{what} is wrong. 
This implies that cognitive processes are involved that try to connect or interpret the perceptional impression with the experienced context or rational analysis. 
For our modeling approach this implies that different levels of control need to be distinguished, along with their specific contribution to the sense(s) of control. 
We will in particular distinguish between two layers in charge of (lower-level) sensorimotor control and (higher-level) cognitive control, respectively. 
Accordingly, we assume a \textit{higher-level SoC} component at the cognitive layer, reflected in (self-)beliefs about the status of one's cognitive action control, and a \textit{lower-level SoC} component residing at the sensorimotor layer, evaluating the success of predictions about an action's effects.

Whether the sense of agency (SoA) can be reduced to an evaluation was discussed by \citep{Chambon:2014cfa}. They differentiate between an experiential SoA that is prospective and thus meta-cognitive, and a judgmental SoA that works retrospectively but lacks perceptual access. In their view automatic processing of action needs to be able to occur without self-attribution judgments and voluntary actions require prospective -- meta-cognitive -- judgments. In contrast, retrospective judgments -- done after the fact -- may or may not require meta-cognitive judgments, depending on the situation.
The function of evaluation that we assume for the sense of control is pre-reflective and is not to be understood as purely meta-cognitive. Rather, it is functional on every level of the proposed hierarchy in the sense that it influences future predictions by weighting how strongly prior information is influenced by new evidence, depending on how accurate the prediction was met. We describe this weighting to be integral to a hierarchy that consists of layers of nested perception-action control loops, where each layer has indirect access to perceptual judgments from the next-lower layer and can be influenced by more abstract judgments from the next-higher layer. This model makes no assumptions about the experience of volitional action nor a necessity of a control influence that is meta-cognitive and integrates judgments of agency, or even a sense of body-ownership. 

Finally, another important question that we want to address with our model is not only how SoC arises during or after situated actions, but also how SoC and its different components in turn affect action control at the two layers.
This involves the important notion of a \textit{predictive SoC}, i.e., the anticipation of the (senses of) control that an agent will have when executing a specific action under specific environmental conditions. 
To address this aspect, we propose the concept of an agent's \textit{action field}, which we define as the space of specific actions it can potentially execute in a given situation. These are represented in terms of the possible effects they can bring about in the environment. SoC is a necessary component of this action field as it directly arises from and, in turn, influences the scope and reliability of one's options for prediction-based action control. We assume that this holds for both low-level and high-level SoC components, at the respective layers and scope of action control. 
{Still, while this work strives to conceptually highlight and shed light on the interdependence and dynamic involvement of -- what we call -- sense of control as well as its underlying causes, we do not see our computational modeling approach as one that enunciates the subjective experience as such.}

In the next section, we describe a conceptual architecture that comprises two layers of control, each of which possessing the ability to evaluate and adapt their respective control strategy to the given situation using prediction-based processes.

\subsection{The free-energy principle}
An increasingly important account of perception and action is based on the \textit{free-energy principle} (FEP), which posits a systemic goal of minimizing so-called free energy, i.e., the uncertainty of a cognitive system in terms of its surprise and divergence with respect to its predictions \citep{Friston:2012fwa}. The FEP's account of action is called \textit{active inference}, which sees action as a form of inference over possible ways to make the environment meet the predictions. The environment is affected through action to reduce uncertainty about predictions that stem from beliefs about the world \citep{Adams:2012gn}. This idea for action control can be traced back to the ideomotor principle describing voluntary actions as resulting from mental representations of their anticipated effects (going back to the analysis by \cite{James:1890wm}).

We adopt this view and assume that low (sensorimotor) and high (cognitive) levels of action control form a generative model, in which predictions about sensory stimuli are continuously formed and evaluated against incoming input across different levels of abstraction. The generative process at each hierarchical layer is then inverted to predict next actions and to attenuate prediction errors \citep{Friston:2010ez}. This mechanism has been discussed as a form of so-called \textit{affordance competition}, a possible mechanism for action selection \citep{Cisek:2007fq} based on possible goals achievable through action.

The success of free energy minimization \citep{Friston:2009iz} is assumed to greatly depend on the balancing of how prediction errors influence prior predictions.
This act of balancing needs to depend on the uncertainty itself, which is described as the \emph{precision weighting} of prior predictions. For example, it has been proposed that the effect of attention -- sometimes inhibiting and sometimes boosting the impact of  sensory data -- is modulated by precision weighting and the surprise of the sensory data \citep{Kok:2012kp,Press:ez}. 


\subsection{Models of action control}
Various approaches have been proposed to model action control in theoretical accounts or in artificial cognitive systems such as robots, control systems or artificial agents. Fewer attempts have been targeted at simulating some form of ``self'' or an epiphenomenal version of what we describe here as a sense of control. A comprehensive model, however, that provides an account of how a sense of control arises from continuous action evaluation within the agent and how this affects the agent's action regulation and decision making is still missing.

One popular approach in (Neuro-)Psychology to explain how an agent evaluates its own behavior is the \textit{comparator model}, basically comparing action goals with the consequential sensory evidence, e.g., in order to distinguish self-caused actions and their outcomes from those caused by others (e.g., \cite{Frith:2000ksa}). This model assumes predictive processes based on inverse and forward models, allowing it to predict and match observed behavior. By limiting its ability to link intended actions to observed behavior it can, e.g., account for disorders of awareness in the motor system and delusion of control. 

The comparator model, however, is mostly concerned with short-term and bottom-up evaluation of behavior, ignoring information from postdictive information processing such as situational context information, intentions or thoughts \citep{Wegner:1999wy}.
It was soon extended to inform a sense of agency that involves top-down and bottom-up processing influenced by more high-level contextual information \citep{Synofzik:2008fl}.
More specifically, the comparator model's sensorimotor information is integrated into a hierarchy to form a non-conceptual level that is described as containing the \textit{feeling} of agency.
Another level of the hierarchy contains a propositional representation, or a \textit{judgement} of agency, that combines contextual or social cues with intentions or thoughts. 
Despite its aspiration for completeness this model makes no claims about how a sense of agency or sense of ownership influences future decision making, although it highlights the importance of the process of being able to compare an intended action to its consequence, and by that evaluate the predictive power of the action's underlying generative model. \\

Work in developmental robotics has looked extensively at how robots can acquire a model of their own actions and the effectiveness of their own body (including a body schema and a sense of body ownership). The most common approach has been to simulate motor babbling, a self-exploring behavior found in infants. Algorithms to mimic this behavior have been implemented in different models to guide the learning of body schemas (or a body map) in different robot platforms. \cite{demirisMotorBabblingHierarchical2005} describe how self-exploration behavior can lead to the learning of useful representations for motor execution and recognition, using a network of coupled inverse and forward models that form a hierarchy of increasingly complex representations. They also describe how a developing body schema allows to develop object-oriented actions, where multiple possible solutions to a problem can co-exist (e.g., different ways of grasping). Likewise, \cite{schillaciPrerequisitesIntuitiveInteraction2011} test different exploration strategies to evaluate their effectiveness in learning a body map, which, as they argue, is crucial for a robot to develop a sense of self. Such accounts inherently afford a plasticity of bodily self representations. However, they do not consider the subjective uncertainty within the agent, and thus the control model's reliability.  


Recent approaches in the FEP literature explore the role and development of a self. For example, \cite{Apps:2014hf} highlight the importance of the ability to represent a self as distinct from others. They propose a theoretical account of self-recognition based on a probabilistic representation of a body-centered version of ``me'', which integrates and interprets (possibly surprising) sensory information in terms of hidden states. It is argued that this probabilistic \textit{me} accounts for psychological evidence that the self and one's ability for self-recognition is malleable to different degrees, e.g., through manipulations similar to that of the rubber hand illusion experiments. In previous work \citep{Kahl:2018ki} described a computational approach of inferring a ``sensorimotor sense of agency'' during behavior production, using a model's free energy as a functional measure of the reliability of the model being responsible for a behavior. This approach links an agent's action to its predicted effect and allows to differentiate it from actions of other agents. In this sense the model infers a minimal (non-experiential) form of ``mineness'' with regard to a perceived behavior. 

We argue that regulating information integration at different levels of action control is vital for adaptive action control in complex and unpredictable scenarios, in which compensatory control at different timescales must be combined with more global changes of control strategies or action goals. Approaches for combining such levels in a unified control hierarchy show this important step, e.g., the HAMMER architecture \citep{demirisMotorBabblingHierarchical2005}. 
However, they still lack the flexibility to evaluate the reliability of a current strategy and integrate this aspect into future decision making, e.g., to switch high-level strategies in situations where the current low-level strategies cannot cope with given problems. 
More recent modeling approaches from Cognitive Neuroscience, such as the ones based on the FEP, are potentially more flexible with respect to the scope of dynamic control for they can evaluate their predictions at every time step by employing generative models that infer nested sequences of state transitions of shorter temporal stretches within sequences of longer temporal stretches \citep{Kiebel:2009fb,Friston:2017fj}.

Many recent approaches in A.I. employ reinforcement learning to autonomously learn control policies that can cope with uncertain environments. However, these approaches suffer from serious scaling issues and in comparison to biological systems are inherently limited in their ability to transfer learned skills to environments different from the one in which training took place. Also, environments that dynamically change pose a serious challenge due to the combinatorial explosion. Consequently, recent work on motor control focused on isolated motor control tasks instead of domains with rich and diverse behavior \citep{merelHierarchicalMotorControl2019}.
\cite{schillingCrystallizedAdaptivityFluid2019} argue for a form of decentralized control to tackle these problems and discuss how biological systems seem to solve these problems by having a \textit{fluid adaptivity} in contrast to only a \textit{crystallized adaptivity} based on previously learned behaviors.
\textit{Hierarchical reinforcement learning} (HRL) proposes a form of decentralized control by learning to operate on different levels of temporal abstraction.
This way, it promises to improve structure exploration, long-term credit assignment and allows to abstract toward representing sequences of lower-level action.
For example, the option-critic architecture introduced the possibility to discover options of behavioral sequences \citep{baconOptionCriticArchitecture2016}.
The ``locomotor'' controller was another important step, as it discovers working controllers and uses them in sequence based on a policy \citep{heessLearningTransferModulated2016}. This allowed to transfer a learned skill by retraining only meta-policies of scheduling the controllers, reducing the amount of training necessary for learning different behavior.
This recent progress on decentralized and flexible approaches led to architectures allowing to play a full game of Starcraft \citep{pangReinforcementLearningFulllength2019} or the ATARI 2600 game \textit{Montezuma’s Revenge}, i.e., they discover subgoals and skills using novel unsupervised and model-free methods without requiring a model of the environment \citep{rafatiLearningRepresentationsModelFree2019}. Likewise, \cite{10.3389/frobt.2019.00123} combined symbolic action planning with reinforcement learning methods to learn robot control under noisy conditions.

In sum, despite impressive advances in robotics and A.I. the mentioned requirements are hardly solved and especially the vast amount of training necessary for many contemporary techniques (based on HRL) poses obstacles that still require substantial progress, e.g., in transfer learning. In particular, current approaches do not meet the challenge of combining flexible and fast adaptability with goal-directed planning and decision-making in uncertain task environments, where \emph{both} processes are informed by self-centered evaluations of the robustness and suitability of their respective action control strategies in the current situation. We argue that such a model would strongly benefit from an introspective measure of confidence in its abilities for coping with the situational demands. This measure needs to be rooted in continuous evaluations of its specific action and thus is more akin to the  \textit{fluid adaptivity} of biological systems. In the present work we focus on the sense of control (SoC) as one of these measures and as a means of weighting information necessary for decision-making in high-level cognitive and low-level sensorimotor control.


\section{Proposed architecture for action control} \label{sec:model}
As discussed in the previous section, modeling action control in uncertain environments involves integrating several levels of prediction-based action regulation (from sensorimotor compensation to decision-making and planning). We argue that such a model benefits greatly
from including a sense of control that integrates the agent's evaluations of its effective and predicted degree of control across these layers. In this section, we propose a computational modeling account for this. We start by laying out a conceptual architecture 
that can integrate information over different layers of a control hierarchy. 

\begin{figure}
\centering
\includegraphics[width=0.7\linewidth]{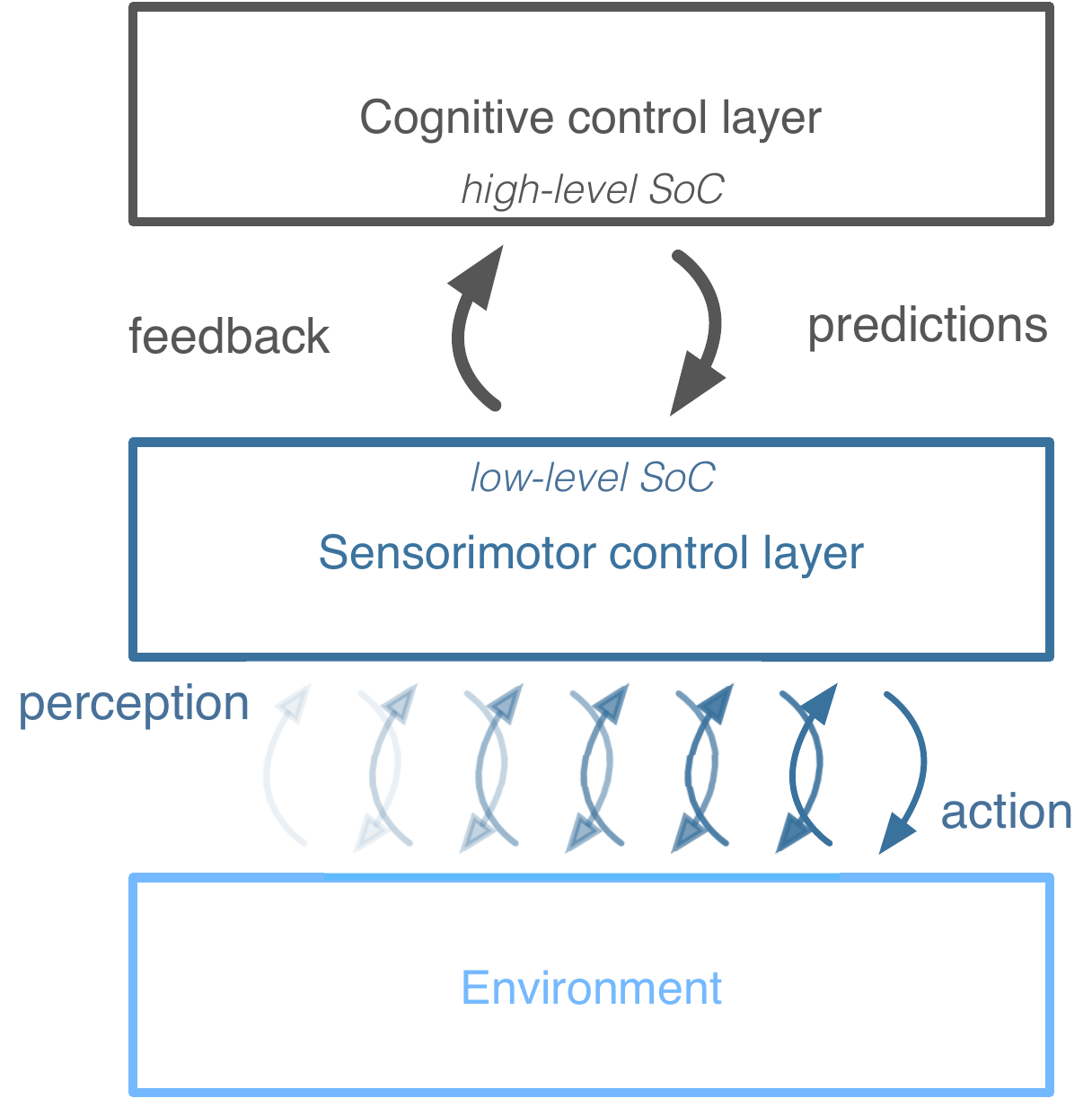}
\caption{The proposed control architecture consists of two layers. It combines top-down symbolic (long-term decision-making) in the CCL with bottom-up sub-symbolic (sensorimotor adaptation) processes in the SCL. The latter acts as the interface with the model's environment. Both layers exchange information using predictions about best movement goals and sensorimotor feedback. The CCL acts on a slower loop with more contextual information, while the SCL acts on a very fast loop with only little information. Both evaluate their own behavior generation and form a sense of control (SoC) -- \textit{low-level} SoC in the SCL and \textit{high-level} SoC in the CCL (which involves low-level SoC).}
\label{fig:model_concept}
\end{figure}

Our modeling approach rests on a number assumptions. First, we assume that action recognition and control rest on principles of \textit{predictive processing} \citep{Clark:2013jo}, in that they minimize uncertainty in action and perception in an integrated manner. We adopt the \textit{free-energy principle} (FEP) to account for this prediction-based, free energy minimizing action, called active inference.

Second, as pointed out above, we assume that continuous control of situated action, with a decisive role of a sense of control, requires an integration of two layers of control. The term \textit{layer} here is supposed to highlight the hierarchical nature of processing and limits the scope of the presented work to a level of description that is more algorithmic and computational, rather than implementational. Layers, here, are understood as containing processes of specific computations that themselves can be of a hierarchical nature. 
We propose an architecture that consists of a hierarchy of two layers (see figure \ref{fig:model_concept}): the top layer called \textit{cognitive control} (CCL) maps cognitive processes that allow for long-term decision making within broad and situated contexts. 
The bottom layer called \textit{sensorimotor control} (SCL) comprises sensorimotor processes that operate on a shorter timescale and ground the model's cognitive processing in the situated environment. Both layers, albeit encapsulating different kinds of processes, together form a generative model in which predictions about sensory stimuli are continuously formed and evaluated against incoming sensory input.

Third, we assume that sense of control arises at both layers and mediates their interplay. In particular, we assume that a low-level SoC reflects the agent's ability to cope with the situational demands through fast sensorimotor adaptation and compensation. This requires a model of how SoC emerges from prediction-based FEP-based sensorimotor processing. Further, a high-level SoC is assumed to reflect the agent's cognitive self-assessment of the control situation. This SoC needs to combine lower-level evaluations with higher-level assessment of, e.g., goal-attainment or plan success, and it provides a basis for the agent's decision-making about strategic adaptations.
In the following we describe the two layers with the respective control processes and sense of control, and how a precision-weighted update mechanism allows the architecture to decide when to stick to a chosen strategy and when to look for better-fitting solutions.


\subsection{Cognitive control layer}
The main task of the cognitive control layer (CCL) is to oversee the whole situation. 
For this purpose, it generates predictions about the current and future circumstances.
It integrates information from the sensorimotor control layer (SCL) and thereby forms an integrated mental model of the situation. 
This model influences the SCL via predictions and in this way orchestrates behavior. 

Central to the CCL is a representation of the agent's \textit{action field}. The action field contains action intentions that previously have been found to be applicable to the current situation. These action intentions are descriptions of a future (or desired) state of the object of reference in relation to its environment. Most of the time this object is the agent's very own body (but can also be an artificial object) and the desired state of this object can be a new location in its environment. 

The actual means -- how to achieve a goal -- are not considered here and left to be figured out by the SCL. This frees the CCL from details that make a situation look vastly different from another. Without this abstraction, it is difficult to represent complex situations such that similarities are recognized that help finding useful action intentions. Hence, finding action intentions that match certain requirements –- we call these the current \textit{situation} -- needs less details in the action representation as well as the representation of the situation. 

The CCL might indeed represent most of the information symbolically, which simplifies the mapping of action intentions to situations.
But there is one problem with symbolic representations in this context: the integration of sensory feedback and the inference of the reliability of the underlying model usually requires a numerical value. As a means to bridge this gap, we propose that the inference of a fitting action intention can make use of a threshold function that chooses situation-intention pairs based on a reliability value. A more sophisticated version of this mechanism may try to differentiate which symbols contribute most to the reliability value. 

Symbolic representations make it easier to model an agent's action intentions and situations by a human designer, introducing the possibility to include past knowledge and to have a reasoning process about the knowledge required to succeed in a given task. This also limits the effect of unintended behavior from learning algorithms, allowing the designer to state explicit rules.

We propose that the CCL could be implemented using the ACT-R cognitive architecture \citep{Anderson2007}, which combines symbolic and subsymbolic processing.
This allows us to leverage a well-established set of theories and pre-existing modules. 
ACT-R by itself is not based on the predictive processing paradigm. 
In ACT-R, functionality that can be executed in parallel is encapsulated in modules. 
Furthermore, the expandability of ACT-R allows us to replace existing modules with ones that better fit our framework. One example is the SEEV vision module \citep{Wiese2019}, which works with abstract predictions about the visual environment and not with concrete instructions. Modules can process information in parallel, but the procedural module as central processing component is serial. This means that only one production (if-then rule) can be processed at one time unit (usually 50ms). Information processed by modules can only be accessed by production rules through buffers, which serve as the interfaces between modules and the procedural component.

\subsection{Sensorimotor control layer}\label{secSCL}
The sensorimotor control layer (SCL) provides three functions in the interaction of the overall model with the external world by means of movement.
The most important is the basic \emph{dynamic control} of movement as a means to fulfill a movement goal through action.
Since the state of the external world is uncertain and can change during an attempted action, or unforeseen forces could hinder or deviate movements and elicit unforeseen action consequences, sensorimotor control has a second function: 
the dynamic \emph{compensation} of deviations from predicted action consequences. 
An assumption of such a compensation mechanism is that there is a force that influences the movement in a way that can then be inferred from the deviation to the predicted action's consequence.
Such an inferred deviating force can be used to dynamically compensate future actions, to the end of minimizing the distance to the movement goal.
Deviations such as these can influence the movement control in an unforeseen way and disturb the sense that you are able to reliably predict the consequences of your actions, i.e., your SoC.
Thus, the third function embedded in the SCL is to provide a SoC based on the ability to reach the current movement goal and to compensate for control deviations.
In the context of the whole model this is a form of \textit{low-level SoC}, limited by the short-term nature of the movement goals and the contextual information that is available.

We propose that the SCL could be implemented similarly to the motor control of the \textit{Hierarchical Predictive Belief Update} (HPBU) architecture \citep{kahlSocialMotoricsPredictive2020,Kahl:2018ki}.
There, at every time step the success of current predictions is evaluated against the sensory evidence at each level of the increasingly abstract sensorimotor hierarchy.
We associate this account of control with active inference because in contrast to traditional control schemes ``[a]ctive inference eschews the hard inverse problem by replacing optimal control signals that specify muscle movements (in an intrinsic frame) with prior beliefs about limb trajectories (in an extrinsic frame)'' \citep[pp. 491]{Friston:2011dw}.
In HPBU and in the model presented here, motor coordination is modeled similarly to \cite{Friston:2011cua}, who applied a dynamical systems approach to motor coordination with specific attractor dynamics for a small handwriting example that allows the motor plant to control a number of reflex arcs and converge on an attractor.
As discussed, in the FEP perspective action is seen as a problem of inference over possible ways to make the environment meet the predictions, i.e., an affordance competition \citep{Cisek:2007fq} for action selection, to the end of reducing uncertainty \citep{Adams:2012gn}.
Such a process circumvents the need for detailed programming of motor commands.
The implementation in SCL (similar to the lower levels of HPBU) is defined as a damped spring system, where the spring's point of equilibrium is set to the action goal's relative coordinates.
This is related to the equilibrium point hypothesis \citep{Feldman:1995ia}, where the predicted sensory consequences of a movement are regarded as the set point to which the motor control plant converges (for more detail please see \cite{Kahl:2018ki} sec. 3.2).

The compensation of noise is in general a mark of sensorimotor coordination and is an integral property of optimal feedback control, where compensation occurs only in variables relevant to maintain the control task performance \citep{Todorov:2002bf}.

We here follow this analysis and describe compensation as a vital part of motor control, in that it corrects prediction errors that are vital to reach the predicted movement consequence.

More specifically, we define prediction errors as deviations from movement consequences as predicted by the motor plant. These are compensated for by systematically changing the movement goal accordingly.
This compensatory adaptation is influenced by cognitive control.
Thus, if a movement's consequence deviates too far from its predicted consequence SCL's own capabilities will not suffice to account for this deviation. It will require help from CCL through the adaptation of the predictions for the motor control plant, depending on the context available in the broader scope of the situation at hand. Accordingly, the proposed architecture's control scheme can be described as consisting of layers of nested perception-action control loops that each work prospectively to reduce uncertainty by means of an affordance competition over their respective state space.


\subsection{Inter-layer information integration}

\begin{figure}
\centering
\includegraphics[width=0.7\linewidth]{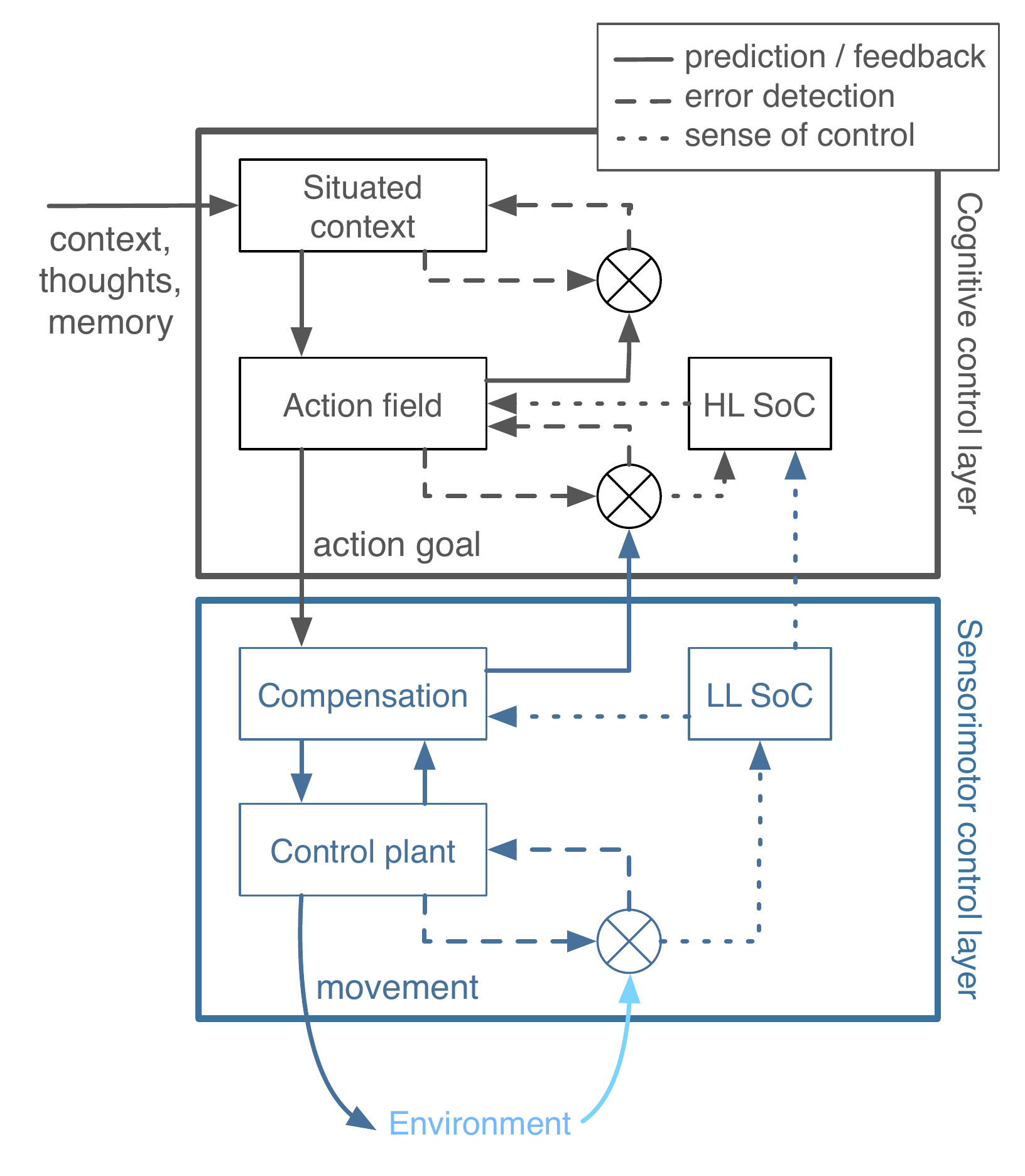}
\caption{The information processing within and between layers of our conceptual architecture. In the CCL, the situated context influences the available action field, which chooses to set an action goal. The action goal influences the sensorimotor control layer's compensation mechanism which translates the action goal to the control plant that triggers movement. Each layer is able to detect and process prediction errors that inform LL SoC and HL SoC and influence aspects of sensorimotor and cognitive control.}
\label{fig:model_interaction}
\end{figure}

The CCL acts on a slower timescale with more contextual information, while the SCL acts on a very fast timescale with only local information. We propose both layers to interact during action control as illustrated in figure \ref{fig:model_interaction}.

Both layers infer their own SoC, incorporating their own perspective on the evidence for successful behavior generation. That is, the CCL infers a \textit{high-level (HL)} SoC with a slowly changing long-term scope, while the SCL infers a \textit{low-level (LL)} SoC with a quickly changing short-term scope. 
The exchange of information between the two layers uses a global mechanism of Bayesian belief updating to the end of minimizing uncertainty (or free energy in the FEP). 
Specifically, we propose a general update strategy, similar to the one presented in HPBU \citep{kahlSocialMotoricsPredictive2020}. There, a linear dynamic update in the form of a Kalman filter is combined with empirical Bayesian updates to integrate information in a hierarchical model that represents its state space discretely.
Each layer is being updated by information it receives from the layer above it, in the form of \emph{top-down} predictive influences as well as from the layer below it, by receiving \emph{bottom-up} evidence. To infer the approximate maximum posterior to decide upon future predictions and action, prediction errors are calculated from this information and are integrated using a linear dynamic update. This update presents the discussed \textit{precision weighting} mechanism where the influence of the prediction error is weighted by the system's uncertainty, in the form of the Kalman gain (here $K$) as a function of the layer's free energy and precision as the inverse variance of prediction error.

More specifically, the top-down information coming from the CCL can contain a predicted action goal, which is set to guide the SCL's actions. The top-down information can also contain further contextual information, if necessary for the layer below to fulfill its goal. The feedback information send bottom-up from the SCL, contains sensory information about the actual state of the control process as well as the LL SoC value as inferred from evaluating the predicted with the actual state of control.

Also, information sent from the CCL and received at the SCL can help to attenuate prediction errors by allowing to evaluate the current situation in the light of a different context. Such top-down predictions can also translate into means to spawn action by way of active inference. Yet, these predictions are integrated into the state of the SCL and can become not only action but also influence and thus help to optimize the action within the broader situated context.
Thus, this combined modeling framework can trigger a form of action which, similarly to the ideomotor principle, creates goal-directed movement by setting the goal.
This is a form of active inference, i.e., making the world conform to your predictions through action.

Thus, in our modeling we assume that movement control does not require centralized planning of every step of the movement, but rather is a combination of a contextually informed goal setting, in combination with decentralized and dynamically adapted control that is informed sensory information.

\subsubsection{Bayesian belief updating strategy}

This is a conceptual model where the details may vary with regard to its application. Generally, each layer $L_j$ represents a discrete number of abstractions over movement consequences. Also, each layer defines a probability distribution $P_t(s_j) \forall s_j \in L_j$ over its discrete domain at each time step $t$ (abbreviated $P^j_t$ for readability). More specifically, SCL would represent probabilities of movement consequences over small deviations from the agent's current position, while CCL would represent probabilities of situated control consequences on a broader spatial or temporal scope, like whole control sequences, abstracted over movements in SCL.
Each layer integrates predictions from more abstract layers and observations from more concrete layers with own latent states. For example, SCL receives predictive likelihoods from CCL $P_t(L_j|L_{j+})$ that are combined in a Bayesian update with SCL's posterior from the last time step as an empirical prior $P^j_{t-1} := P_{t-1}(L_j)$, resulting in a top-down posterior $P_t^\text{j+}$. Similarly, a bottom-up posterior $P_t^\text{j-}$ is updated from a likelihood over movement observations $P_t(L_j|L_{j-})$ and SCL's empirical prior.
Both SCL and the CCL integrate bottom-up and top-down information handled by a belief-update function that combines empirical Bayesian with a linear dynamic update.
At each layer, the update integrates the probabilities of predictive top-down information $P_t^\text{j+}$ from a next-higher layer with probabilities of bottom-up sensory evidence $P_t^\text{j-}$ from a next-lower layer to update the posterior probability distribution $P_t^j$.

This update is similar to the belief update strategy applied in \cite{kahlSocialMotoricsPredictive2020}.
A Kalman filter is used with a layer-specific Kalman gain $K_t^j$ as a function of free energy $F_t^j$ and precision $\pi_t^j$, i.e., the inverse variance of the prediction error ($P_t^\text{j-} - P_t^\text{j+}$):
\begin{align}
P_t^j &= P_t^\text{j+} + K_t^j (P_t^\text{j-} - P_t^\text{j+}) \label{eq:belief_update} \\
K_t^j &= \frac{F_t^j}{F_t^j + \pi_t^j} \label{eq:kalman_gain} \\
\pi_t^j &= \text{ln}\frac{1}{\sigma^2(P_t^\text{j-} - P_t^\text{j+})} \label{eq:precision} \\
F_t^j &= H(P_t^\text{j+}) + D_\text{KL}(P_t^\text{j+}||P_t^\text{j-}) \label{eq:free_energy}
\end{align}
Eq.~\ref{eq:free_energy} models the free energy $F_t^j$ at layer $j$ as a function of the entropy $H(P_t^{j+})$ of the prediction distribution and the cross-entropy $D_\text{KL}(P_t^{j+}||P_t^{j-})$ between the prediction and the evidence.
This belief update strategy integrates the filter (eq.~\ref{eq:belief_update}) into the \emph{dynamically} updated hierarchical model context.
The resulting new posterior $P_t^j$ is also send to the next-higher and next-lower layer to inform their updates as evidence or as a prediction, respectively.
A decision is made by selecting the approximate maximum posterior from $P_t^j$. The associated representation can either be used to predict sequences of states or single states in the next lower layer.


\subsection{Modeling and integrating sense of control}
The sense of control is assumed to form through the continuous evaluation of goal-directed actions and their effect on the environment.
In the proposed control architecture (see figure \ref{fig:model_interaction}), SoC at the different layers influences the dynamic interaction between the perceived evidence and the predictions based in the different layers of abstraction, by influencing the scope of control and by that the ability to choose a fitting control strategy for the situation.

In the following, concepts for inferring SoC and its effect at the respective layers of the control architecture are described. Also, we describe details of a first concrete implementation of the architecture in a specific task scenario (described in the next section).

\subsubsection{High-level sense of control}
At the CCL data about previous goals and parts of the \textit{active self}, the representation of the current situation, are used to find goal-directed actions.
At this layer, it is not important how exactly these actions are performed, but to infer from the feedback information hypothesis about which aspects of the environment are relevant to the action outcome. 
For this matter, the CCL monitors the efficacy of predicted \textit{action goals} and the provided \textit{LL SoC}, given certain situations (see top half of figure \ref{fig:model_interaction}). 
The result is a HL SoC, which acts as a gatekeeper of the situation model updating process. 

The CCL receives information from the SCL which it can evaluate, finding the difference between the discretized sensorimotor feedback and the currently predicted goal. If the LL SoC is too low over a longer period of time HL SoC will also decrease. If HL SoC is below a certain threshold, it triggers the selection of a new action from CCL's \textit{action field} (see figure \ref{fig:model_interaction}). To make an informed decision, the model detects the current situation and looks for adequate actions in its action field (action selection forms the action field from a number of possible actions). The threshold that triggers the selection of a new action or strategy is the \textit{CCL Threshold}, which in our current model iteration is fixed to a certain value.  

The more often actions fail to be successful the weaker the HL SoC will be in a given time frame. If the action field is depleted and the situation has not changed, HL SoC would have to be reduced significantly, as no further strategy for action is available. What if the CCL (constantly monitoring its action efficacy) would only find ineffective actions? We propose two possible solutions: \textit{(1)} the action field only consists of actions the execution of which would further decrease the LL SoC, but because the situation is not changing significantly, still the next viable action from the action field is selected. Once this action field is depleted, simply a new action field is generated that consist of the same actions, resulting in an endless loop of trying the same actions over and over again without success in increasing LL SoC.

This leads to another possibility: \textit{(2)} with a depleted action field there are no actions left to try out and simply regenerating the action field does not yield any new actions. In effect this situation dramatically reduces the HL SoC. Such a situation, with HL SoC falling below a certain threshold, can allow the CCL to readjust its evaluation function to compute the viability of action goals. 
The resulting re-evaluation of possible action goals can allow for taking other action goals into consideration that previously were not perceived to be viable options. 
We think of this as resembling exploratory behavior, e.g., from trying to look at a problem from a different perspective, or putting yourself into another person's shoes, down to forms of motor babbling to explore new action possibilities.

In the current model iteration the first solution is implemented and tested, where simply a new action field is generated once it is depleted.  


\begin{figure*}[ht!]
\centering
\includegraphics[width=0.9\linewidth]{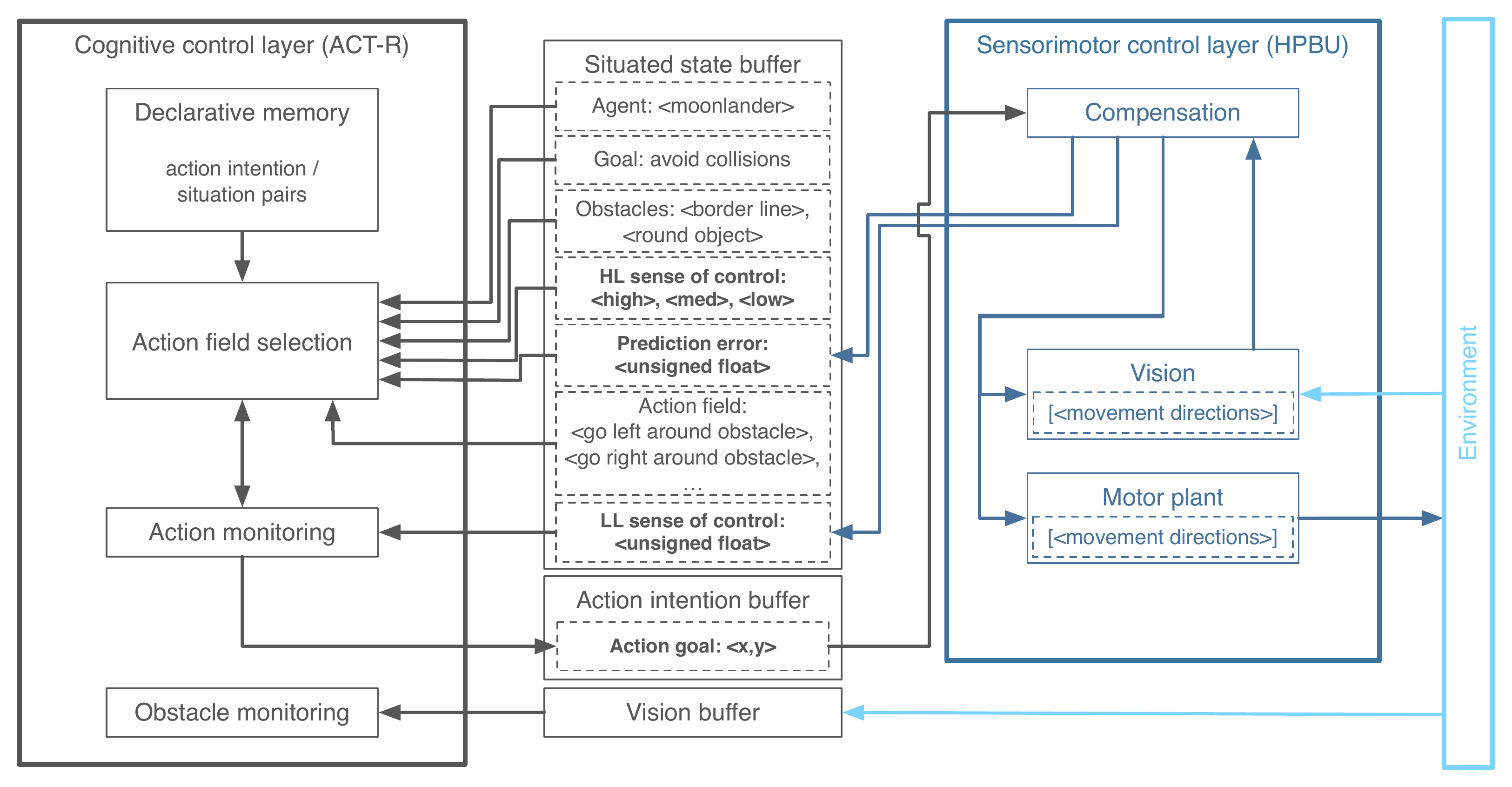}
\caption{Influence between SCL and CCL is realized via the situated state buffer where information from both layers are stored. CCL can use its higher-level reasoning about the current situation and link it with contents of the situated state buffer. CCL's slower loop of reasoning can influence the faster loop that is implemented by SCL. There, less but more specialized information about the next movement goal is used to produce movement and to evaluate that movement. Both layers infer their own control evaluation in a LL SoC and HL SoC, which in turn influences the selection of action strategies from the action field.}
\label{fig:architecture_implementation}
\end{figure*}

\subsubsection{Low-level sense of control}
The LL SoC in the SCL is modeled similarly to the approach presented by \cite{Kahl:2018ki}, which encompasses not only the success of prior predictions of control movements.
Movement is dynamic, so it is unsurprising that \cite{Chambon:2014cfa} found a fluency effect in sense of control, in which the repeated success of predicting and selecting actions seems to increase it over time.
We propose that this is realized using a Kalman filter for the integration of evidence for the success of sensorimotor control over time, i.e., the evaluated predictions of movements consequence and timing (for more detail see figure \ref{fig:architecture_implementation}).
In that the success of the architecture's ability to minimize free energy is evaluated and hence future decision making about the selection of action goals and movements is influenced through the described process of belief updating using empirical Bayesian and linear dynamic updating that is weighted using the model's uncertainty (based on free energy and precision).

Together, the proposed architecture provides a functional form of an ``active self'' by forming a continuous sense of control and using it actively for flexible action control. This includes adaptation through short-term compensation in the SCL, as well as allowing for information from short-term control to influence the planning and decision-making in the longer term. Long-term decision-making in the CCL is also influenced by situated context information and goal setting. Processes in both layers are informed by the evaluations of their robustness and suitability in the current situation through processes that allow to assess the respective layer's prediction errors.

\subsubsection{Implementation of the control architecture with ACT-R and HPBU}

To provide simulation-based evaluations of the conceptual control architecture we implemented it specifically to the control task scenario (described in section \ref{sec:simulations}). It uses ACT-R for the cognitive control layer. For SCL we adopted and extended HPBU's control layer for the representations required by the control task (as described above in section \ref{secSCL}). SCL represents movements and their strength in only two directions, while CCL represents control goals that depend on their availability in the action field.
The communication and influence between both control layers is depicted in figure \ref{fig:architecture_implementation} and is described in the following: The production system of ACT-R (if-then rules) allows to easily model the symbolic and subsymbolic processing of the cognitive control layer. 
Inter-layer information integration is realized by the buffer interface. The \textit{situated context} (see figure \ref{fig:model_interaction}) is represented by the \textit{situated state buffer}. Information from both layers are stored in this buffer. 
The cognitive control layer can reason about the current situation by linking the contents of the situated state buffer with memory chunks from declarative memory. ACT-R allows to compile this knowledge into procedural knowledge. This is the previously mentioned slower loop of reasoning and learning. Whereas the faster loop is implemented in the sensorimotor control layer and uses less but more specialized information.

\textit{Action field selection} chooses from a number of possible actions that form the action field. Action intention/situation pairs are retrieved from declarative memory and checked against a reliability threshold (currently predetermined). It allows the agent to use only actions with a reliability higher than that. Situation representations include the goal and the agent.

The estimation of SoC $\hat{s}_t$ for the current time step is done in both layers of the two-layer hierarchy: for CCL as a high-level SoC (\textit{HL SoC}) $\hat{s}^{hl}_t$, and for SCL as a low-level SoC (\textit{LL SoC}) $\hat{s}^{ll}_t$.

\textit{LL SoC} $\hat{s}^{ll}_t$ is estimated in the \textit{compensation} component of the SCL by a Kalman filter from the Gaussian likelihood of the intended movement and the previous LL SoC estimate: 
\begin{align}
    \hat{s}^{ll}_t &= \hat{s}^{ll}_{t-1} + K_t~(p(m^\prime) - \hat{s}^{ll}_{t-1}) \\
    p(x) &= \frac{1}{\sqrt{2\pi\sigma}} ~exp\left\{{-\frac{(x-\mu)^2}{2\sigma^2}}\right\}
\end{align}

The Gaussian likelihood of the movement describes the difference between the currently perceived movement $m^\prime$ and the intended (or predicted) movement $\mu=m_I$.
As for the precision-weighted belief update, Kalman gain $K_t$ integrates the filter into the model's context of uncertainty.

The HL SoC is updated step-wise in ACT-R every 50ms:
\begin{align}
    \hat{s}^{hl}_t = \hat{s}^{hl}_{t-1} + \left\{ 
    \begin{aligned}
        \tfrac{1}{10} & ~ ~\textit{if} ~ ~\hat{s}^{ll}_t ~> ~ \hat{s}^{hl}_{t-1} \\
        -\tfrac{1}{10} & ~ ~\textit{if} ~ ~\hat{s}^{ll}_t ~< ~ \hat{s}^{hl}_{t-1} - \tfrac{1}{10}
    \end{aligned}\right.
\end{align}
This allows HL SoC to update slower than LL SoC, becoming less volatile. Its updating in $\frac{1}{10}$ steps discretizes LL SoC, which may help us with future symbolic processing in ACT-R.

\textit{Action monitoring} uses the \textit{HL SoC} to assess the effectiveness of the current action. Below a threshold (CCL-Threshold), a new action intention from the action field will be selected. Action intentions are represented symbolically and are translated into action goals using a specialized production. To keep the new action goal from switching again during the first moments, where the new action goal is evaluated in LL SoC and HL SoC, a 300ms window is granted in which action monitoring is halted.


\section{Simulation and results}\label{sec:simulations}

\begin{figure*}[ht!]
\centering
\includegraphics[width=0.9\linewidth]{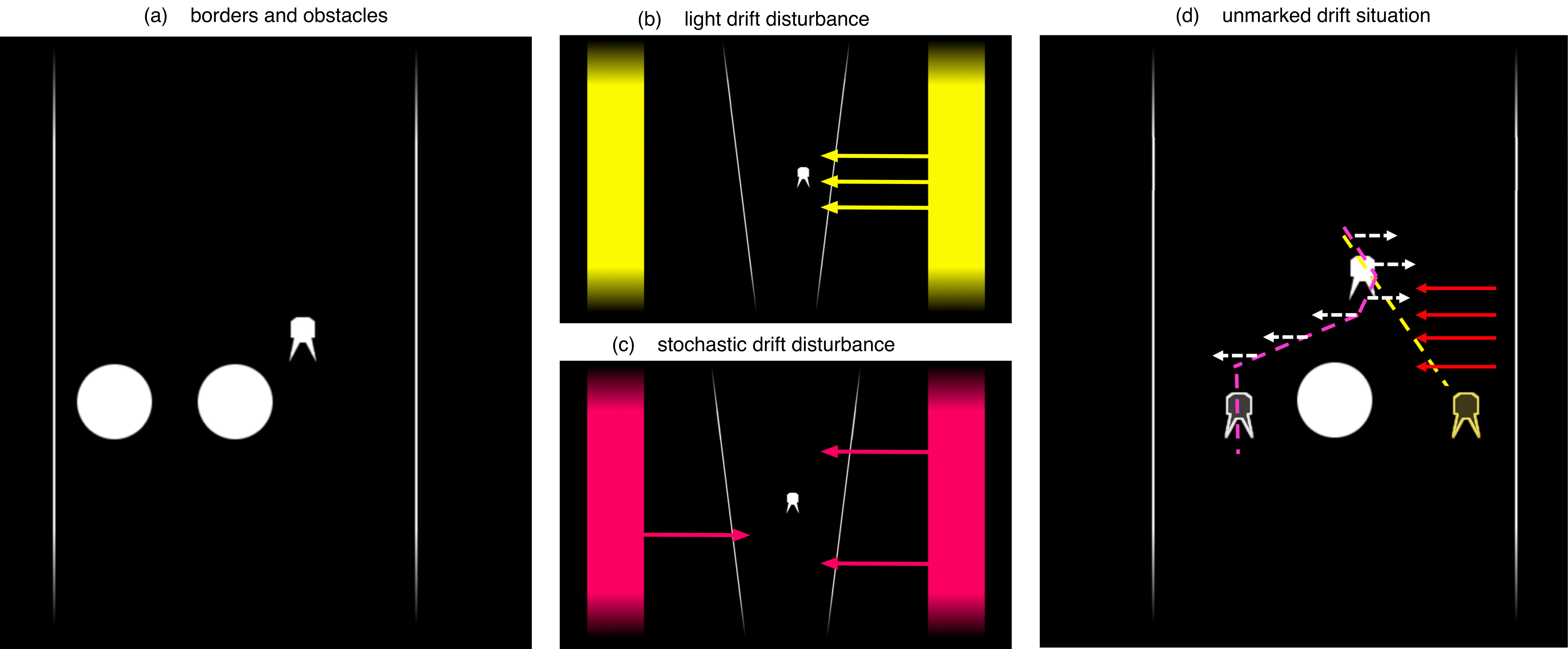}
\caption{Examples from the Moonlander task where a spaceship has to be controlled safely. \textit{(a)} shows the ship in the center, next to two obstacles and the borders left and right to avoid. \textit{(b)} shows a light drift disturbance marked in yellow. \textit{(c)} shows a stochastic drift disturbance and \textit{(d)} an unmarked drift situation where compensation can be difficult without crashing.}
\label{fig:compas_event_examples}
\end{figure*}

In order to develop, apply, and test the proposed control architecture, we devised a task scenario for goal-driven situated action in unpredictable environments. The following scenario was designed to serve both, as setting for empirical investigations of corresponding human behavior in this task as well as a simulation test bed for evaluating implemented \emph{computational models} of the proposed architecture. 
The task is designed to require an agent to act in an embodied way in an environment with varying degrees of uncertainty. This is needed to evaluate the following aspects of the artificial cognitive agent: adaptability of the SCL to environmental factors through short-term processes; long-term planning and decision-making at the CCL; the interaction between SoC and a robust representation of the situation and the active self. To be able to study these aspects both with an artificial agent and with human participants, we chose a video game setting similar in style to the classic ``Moonlander'' or ``Lunar Lander'', with game levels that represent different requirements of control difficulty.

In the proposed task an agent has to steer a Moonlander (-style) spaceship in order to avoid crashes with level-borders or obstacles. The game is relatively simple at a first glance: the spaceship descends automatically with a fixed speed. Therefore there are only two possible movement inputs at each step of the simulation: to steer right or steer left, both move the ship immediately in the selected direction. There is no complex or realistic physics simulation, e.g., no inertia or turning radius. 
In addition, disturbances were included that introduce different degrees of drift. These are either predictable through the color of the level borders or stochastic and thus unpredictable. These provide opportunities for adaptation and learning of human participants. For examples of the scenario's design, please see figure \ref{fig:compas_event_examples}. It shows the general level design with borders and obstacles (figure \ref{fig:compas_event_examples}a) as well as examples for predictable drift disturbances (see figure \ref{fig:compas_event_examples}b) and unpredictable stochastic disturbances (see figure \ref{fig:compas_event_examples}c). Also included are unmarked drift disturbances that occur in combination with an obstacle (see figure \ref{fig:compas_event_examples}d). These provide an opportunity to observe the agent's (or participant's) behavior when a set strategy or action goal (yellow spaceship) is no longer viable and should be changed (to the goal of the white spaceship).

\subsection{Evaluation}

We ran simulations of the implemented model with different setups to test both, its ability to avoid crashes and to see the influence of its decision making on its HL and LL SoC. In addition, we measure the number of observable \textit{strategy changes} since it depends on both HL and LL SoC and could easily be compared with empirical results from human participants.
We define a strategy change to be a significant directional change that is then kept for over at least 6 simulation steps. 
Also, we analyze the model's SoC prior (150ms) and posterior (600ms) to a strategy change.
We control two variables in our simulations: one is the Kalman gain in SCL's compensation component. It weighs the influence of CCL's compensation strategies against the automatic compensation that reacts to bottom-up evidence and also it weighs the strength of updating LL SoC. The other variable is the previously mentioned CCL Threshold that triggers a cognitive process for a strategy change when the HL SoC drops below this value. 

As a baseline, the simulation was run with only the SCL enabled and a dynamic Kalman gain (K)  (see equation \ref{eq:kalman_gain}). Then, the complete control hierarchy was enabled so as to allow for strategic control at CCL to have an effect. Here, the precision weighting between cognitive control and sensorimotor control was balanced to allow for future movements to be equally influenced by strategic and reactive compensation (with a fixed instead of a dynamic $K=0.5$). Then again the complete model hierarchy was tested, but with precision weighting between cognitive control and sensorimotor control set to be biased toward cognitive control influence (with a fixed $K=0.3$). Finally, the same setup was used but with precision weighting between cognitive control and sensorimotor control set to be biased toward sensorimotor control influence (with a fixed $K=0.7$). The complete hierarchy simulation conditions were repeatedly run with different CCL Thresholds ($0.3$, $0.5$, and $0.7$).

Each setup condition is simulated for six contained level situations. This allows us to measure the impact of specific situations, without having to take into account the potential influence of previous situations. Level situations are (a) \textit{cone with marked light disturbance}, (b) \textit{no disturbance, but obstacles at left and center}, (c) \textit{cone with stochastic disturbance}, (d) \textit{unmarked medium disturbance with a long reaction until an obstacle}, (e) \textit{unmarked medium disturbance with a short reaction time until an obstacle}, and (f) \textit{unmarked light disturbance with a long reaction time until an obstacle}.

\subsection{Results}

\begin{figure*}[ht!]
\centering
\includegraphics[width=\linewidth]{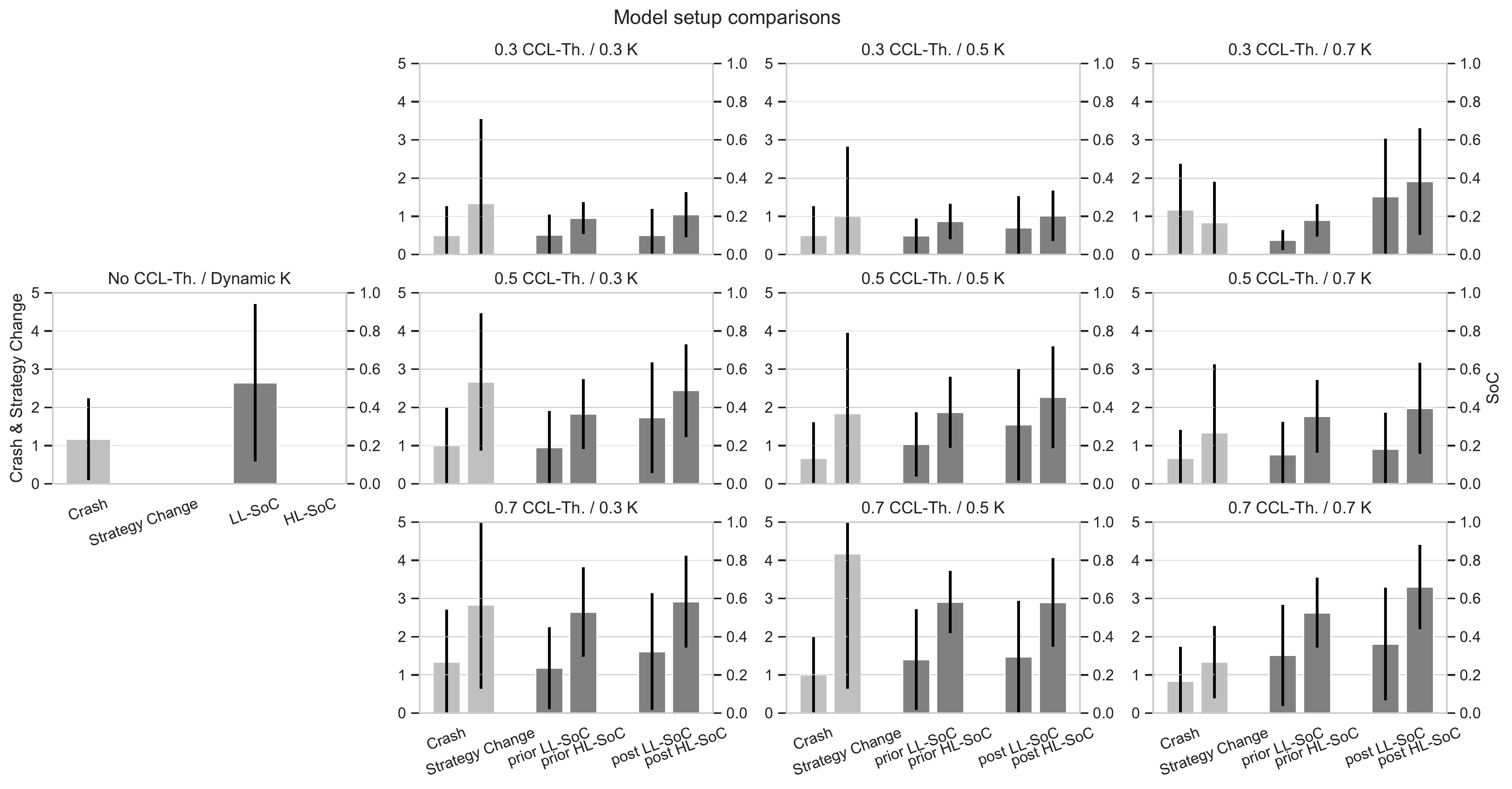}
\caption{The simulations evaluate the model's performance in different setups. On the left, the resulting average number of crashes and LL SoC over all six level situations is shown for the baseline simulation with only the SCL layer enabled and using a dynamically calculated Kalman gain (K). The different CCL-Threshold and K combination simulations have SCL and CCL enabled. We show the resulting number of crashes and strategy changes, as well as the LL SoC and HL SoC prior and posterior to strategy changes averaged over all six level situations. }
\label{fig:simulation_results}
\end{figure*}

We ran a total of 60 simulations to evaluate the model's performance: six level situations tested in the baseline setup and all nine setup combinations of K and CCL-Threshold. The left column of figure \ref{fig:simulation_results} shows the resulting average number of crashes and LL SoC over all six level situations for the baseline simulation with only the SCL enabled, which is using a dynamically calculated Kalman gain (K). The other plots show the different CCL Threshold (CCL-Th.) and K combinations, with both SCL and CCL enabled. We show the resulting average number of crashes and strategy changes, as well as the LL SoC and HL SoC prior and posterior to strategy changes over all six level situations.

The LL SoC and HL SoC evaluate the agent's control in the short term and the long(er) term and influence its decision-making. In most of the model configurations, the average HL SoC is higher than the LL SoC, reflecting the effects of continued local adaptations the SCL has to carry out. Moreover, the average SoC in most situations tends to be lower before a strategy change than after it, demonstrating that the SoC facilitates strategy adaptations to action control that help to improve the agent's self-assessed control situation.

Note that a detailed analysis of the effects of different parameter settings, like CCL thresholds and K values, are beyond the scope of this conceptual paper which is meant to introduce and test the overall conceptual approach. Still, qualitatively, an interesting result is that there is no straight-forward formula for avoiding crashes. A high CCL threshold increases the number of strategy changes but does not lead to fewer crashes, although it certainly leads to higher LL SoC and HL SoC overall. A high K increases the weight on integrating bottom-up information which leads to a reduction of strategy changes by increasingly disregarding CCL's compensation strategy. Interestingly, this does not necessarily lead to more crashes. Still, a more balanced $K=0.5$ results in simulations with the fewest number of crashes. Also, higher CCL thresholds seem to increase the number of strategy changes and the overall maximum LL- and HL SoC, while a higher K seems to increase the positive impact of strategy changes on posterior LL- and HL SoC and decreases the number of crashes for higher CCL thresholds.  
Overall, having a CCL that takes a broader scope into account can be beneficial to reduce the number of crashes. On the same note, having SCL be able to evaluate its actions which -- reflected in its LL SoC -- allows for strategy changes to be applied from CCL that leads to increased LL and HL SoC. Generally, strategy changes have a positive impact on LL SoC and HL SoC but do not decrease the number of crashes per se.

\subsection{Example SoC dynamics}
\begin{figure*}[ht!]
\centering
\includegraphics[width=0.7\linewidth]{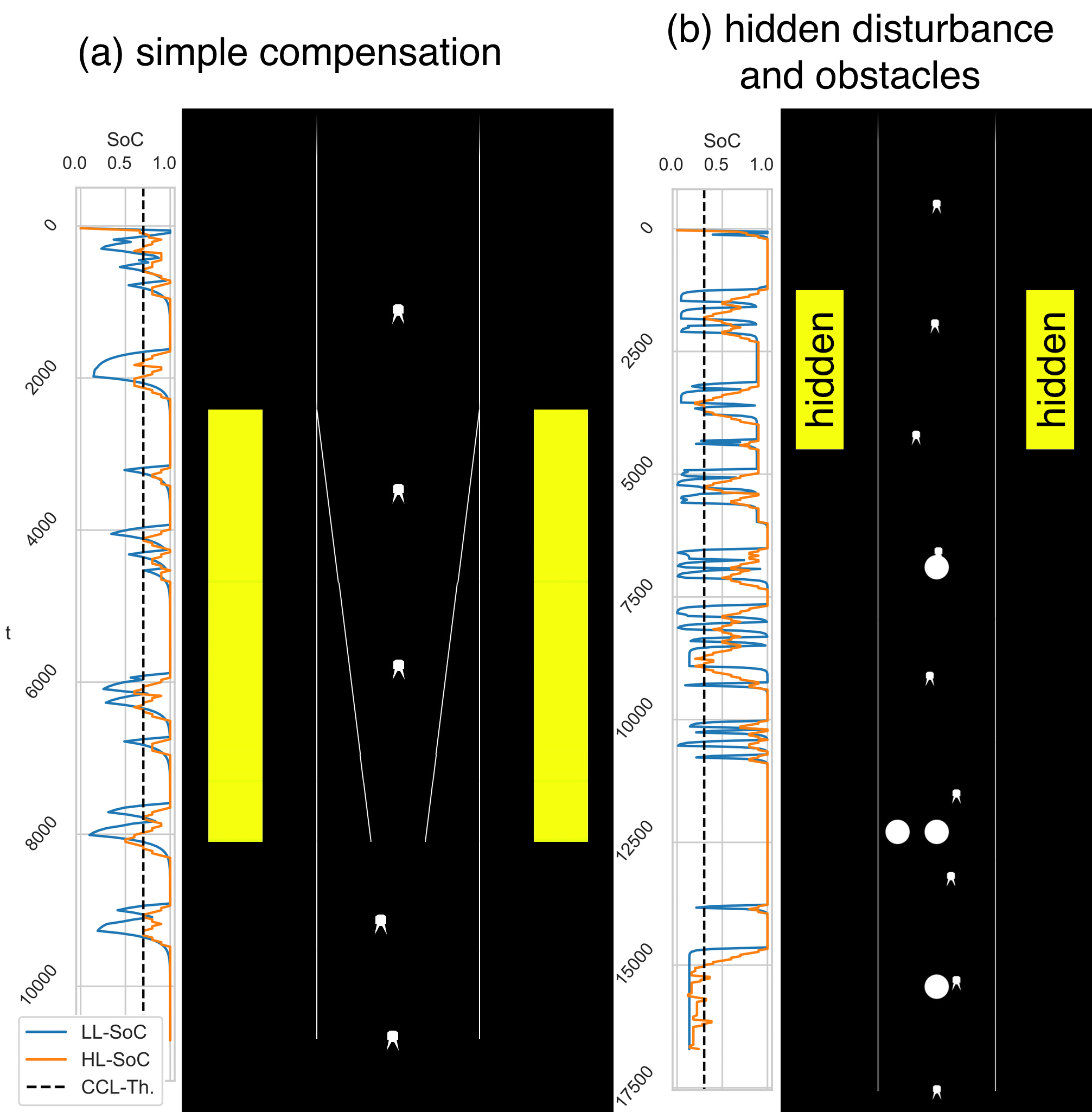}
\caption{Two examples from the Moonlander task scenario that show the SoC dynamics that evaluate the control reflected in LL- and HL-SoC. \textit{(a)} shows a simple compensation example with strategy changes clearly reflected in the drops below the CCL-Threshold. \textit{(b)} shows an example with hidden disturbances, for which only SCL can compensate without strategic influence from CCL, after which it fails to avoid the following obstacle.}
\label{fig:soc_influence_example}
\end{figure*}

To get an impression how the two levels of SoC are affected by their respective layers of control we  take a closer look at two situations of the task scenario. Crashes, disturbances and obstacles influence LL and HL SoC through the dynamics found in SCL's and CCL's ability to cope with the control task. As described, SCL's evaluation of control is reflected in the compensation component's Kalman gain (K) - a function of the component's underlying free energy (F) and precision values - which in the LL SoC update weighs the Gaussian likelihood of the current movement goal (see eq. \ref{eq:kalman_gain}). HL SoC is a higher-level representation of this, evaluating control toward the action goal. Figure \ref{fig:soc_influence_example} shows the dynamics influencing LL and HL SoC evaluations of control in two very different level situations. Figure \ref{fig:soc_influence_example}a) shows a simple compensation example with a color-marked disturbance area that allows CCL to select a compensation bias, influencing SCL. When HL SoC drops below the marked CCL-Threshold a new strategy will be selected. The model can navigate this situation successfully without crashing. Figure \ref{fig:soc_influence_example}b) shows a more difficult example with unmarked hidden disturbances that cannot be used by CCL to select a compensation strategy so SCL has to compensate on its own. Shortly after, an obstacle becomes visible that CCL is not able to find a successful strategy for, which results in a crash. In the following, CCL first wants to move to the middle of the level, just before two obstacles become visible. These can be circumnavigated successfully. For the last obstacle multiple strategies are chosen one after the other, resulting in a crash. Both SoC values do not have a chance to recover from this.

These detailed examples, in addition to the evaluating simulations described above, speak to the possibility of planning action strategies based on the model's current evaluation of its own performance both, on the level of fast but more basic movements and on the level of slower but strategic action. Currently, the threshold for strategy changes in CCL is set by hand. This could be replaced by a more informed decision-making based on the learned reliability of single strategies in the current context (obstacles and marked disturbances) as well as information from SCL.


\section{Conclusions and prospects}
In this paper we have presented a conceptual control architecture that accounts for situated action in partially unpredictable environments and aims to help to understand the embodied cognitive processes that lead to a sense of control as part of the experience of an active self. We believe that the proposed approach represents a novel contribution to the fields of Cognitive Systems and Robotics for several reasons: 

First, it explicates how to integrate two key layers of action control, a higher-level cognitive control layer with a lower-level sensorimotor control layer. While each of these kinds of layers have been studied before in largely separate fields, their integration still poses many open questions especially in goal-driven and robust behavior in unknown and unpredictable situations. We propose an integration based on larger principles of prediction-based processing.

Second, we argue that this integration is indispensable to account for the active self and, more specifically, a sense of control in autonomous agents. We provide a novel account of a sense of control for artificial cognitive systems \citep{ThomazA.L.LievenE.CakmakM.ChaiJ.Y.GarrodS.GrayW.D.LevistonS.C.PaivaA.&Russwinkel2018}, allowing them to continuously assess their options for actions as afforded by the world and using this as part of a situated self-representation that informs goal selection as well as choice or adaptation of control strategies. Our approach embodies a process-oriented account that could provide testable ideas as to how this can be achieved. Its value arises from the fact that it makes theoretical hypotheses and modeling assumptions explicit, puts them to scrutiny with regard to their completeness and level of detail, and makes them falsifiable by providing predictions that can be tested against empirical observations. The control architecture can in that way serve to develop more refined models and to conduct experimental studies on cognitive embodiment phenomena. In that way it will enable researchers from a larger community to test different hypotheses rooted in the theoretical background that the extended architecture offers.

Third, we contribute to the approach of using artificial cognitive agents as a valuable tool to investigate phenomena of subjective experience typical of humans. Specifically, our work enables to test a functional account as to which mechanisms underlie certain aspects of an active self, and to study whether its implementation shows similar behaviors as humans do in difficult dynamic situations. This may yield important insights into the embodied cognitive mechanisms that afford the needed plasticity of the mind’s embodied representations and still enable a robust self-representation to underlie situated action control.
A possible line of investigation is that of inter-personal variability with regard to an individual's subjective experience of action. Different settings of proposed parameters (e.g., the Kalman gain parameter K, here used for precision-weighting) may allow the model to match control behavior similar to a subject's, whose subjective experience could be judged to be more focused on the detail of an action, while others experience it more on a consequential level \citep{vallacherLevelsPersonalAgency1989}.

Fourth, in addition to the proposed control architecture we put forward a task scenario which not only allows for testing humans and artificial agents. It also provides an environment in which agents need to adapt to new situations and thus might learn in an open-ended fashion how to apply sensorimotor adaptation to differing situations with an expected outcome even when conditions change. This may potentially enable learning-based approaches to develop the cognitive and embodied mechanisms needed to experience sense of control and to develop a sense of an artificial self.

The simulation-based evaluations of an implementation of the proposed conceptual architecture highlight the importance of weighting the information integration from both layers of control. Giving too much weight to one over the other can work in some situations but may be detrimental in others. We plan to integrate this information more dynamically in future model iterations. Similarly, the threshold for selecting higher-level action and compensation strategies can work in some situations but may lead to quick strategy changes that are too volatile. We plan to have strategies be associated with a value of reliability that allows to have more informed decision making in this regard. Such an associated reliability value could learned from experience, where the model remembers successful strategies and applies them in similar situated contexts.

Finally, robotic systems equipped with the proposed control architecture could make use of cognitive mechanisms that would enable new ways of collaborative human-robot interaction. This holds for mechanisms such as the ones introduced here when, e.g., a robot may inform its human partner of its low SoC, switch to a safer mode of operating, or ask for help. Further, the proposed architecture provides a basis for other sophisticated mechanisms needed for collaborative situated interaction, like embodied Theory of Mind and mental (co-)simulation \citep{Russwinkel2020,poeppelkopp2018,kahlSocialMotoricsPredictive2020}. For example, the presented model can enable a robot to anticipate not only its own action field, but also another one's actions and their expected outcomes \citep{Klaproth2020_topics}.

We believe that the presented model provides a broad range of opportunities to study the nature of the active self. Our task scenario directly enables empirical studies on human behavior in these situations, along with additional measures of sense of control. We have ran first pilot studies and gathered results on crashes, strategy changes and (self-reported) SoC in human participants. While the results of this analysis are to be reported elsewhere, we are confident that the combination of computational and empirical investigation will generate important and novel findings that will help to elucidate the neural and psychological basis of the active self and the sense of control, more specifically.

\section*{Acknowledgments}
This research/work was supported within the Priority Program ``The Active Self'' funded by the German Research Foundation (DFG).

\section*{Conflict of interest}

The authors declare that they have no conflict of interest.

\bibliography{main.bib}
\bibliographystyle{apalike} 

\end{document}